\documentclass[11pt]{article}

\usepackage[final]{acl}

\usepackage{times}
\usepackage{latexsym}

\usepackage[T1]{fontenc}

\usepackage[utf8]{inputenc}

\usepackage{microtype}

\usepackage{inconsolata}

\usepackage{graphicx}

\usepackage{amssymb}
\usepackage{bm}
\usepackage{amsmath}
\usepackage{subcaption}
\usepackage{booktabs}
\usepackage{xcolor}
\usepackage{placeins}

\newcommand{\stitle}[1]{\noindent\textup{\textbf{#1}}}
\newcommand{\ind}[1]{\mathbf{1}\!\left[#1\right]}

\newcommand{\msep}{\mathbf{m}^{\mathrm{sep}}}
\newcommand{\mctx}{\mathbf{m}^{\mathrm{post}}}
\newcommand{\mrng}{\mathbf{m}^{\mathrm{rng}}}
\newcommand{\mboth}{\msep\!\odot\mctx}

\title{GAPS: Dimension-Level Gates for Conditional Activation Steering}

\author{Moghis Fereidouni \\
  University of Kentucky \\
  \texttt{moghis.fereidouni@uky.edu} \\\And
  Muhammad Umair Haider \\
  University of Kentucky \\
  \texttt{muhammadumairhaider@uky.edu} \\\AND
  Hassan Sajjad \\
  Dalhousie University \\
  \texttt{hsajjad@dal.ca} \\\And
  A.B. Siddique \\
  University of Kentucky \\
  \texttt{ab.siddique@uky.edu} \\}

\begin{document}
\maketitle

\begin{abstract}

Activation steering suppresses undesired behaviors in language models by adding a steering vector to the hidden state during generation. Recent conditional methods such as CAST and DSAS improve the behavior–capability trade-off by deciding when to intervene, but once active, they apply the full dense vector to all hidden dimensions, regardless of whether a neuron carries concept information or already lies in the desired regime. We introduce dimension-level conditioning as a complementary axis of selectivity that also decides which neurons to intervene on. Our method, GAPS (Gated Activation steering via Posterior and Separability), combines two training-free gates: a static separability gate that restricts steering to neurons with statistically reliable concept information (via AUROC), and a dynamic posterior gate that steers a neuron only when its current activation is better explained by the undesired concept under a Gaussian model. The gates add O(D) overhead per token, and they plug into existing conditional methods. On toxicity mitigation (RealToxicityPrompts) and concept removal (OneSeC) with Gemma-3 (4B) and Qwen-3 (1.7B), GAPS consistently matches or improves the Pareto front of its token-level counterparts; under a fixed capability budget, DSAS+GAPS reduces Gemma-3's toxicity rate from 6.52\% to 0.48\%, versus 3.52\% for DSAS alone. Ablations attribute most of the gain to the posterior gate.

\end{abstract}

\section{Introduction}

Language models often need to avoid unwanted outputs, such as toxic continuations or references to a specific concept, but retraining the model for every new constraint is impractical. Activation steering offers a low-cost alternative: a steering vector, typically computed as the mean difference between hidden states collected from examples of desired and undesired behavior, is added to the model’s hidden state during generation \citep{rimsky-etal-2024-steering, suau2024whispering, rodriguez2025controlling, rodriguez2025lineas}. This approach requires no gradient updates and adds little inference cost. However, conventional activation steering applies the same dense vector at every generation step, regardless of whether the current context warrants intervention. Interventions can therefore degrade fluency and downstream accuracy. Steering methods are best compared by the trade-off they achieve between behavior change and capability loss.

Recent work improves this trade-off by deciding \emph{when} to intervene: token-level conditional methods such as CAST \citep{lee2025programming} and DSAS \citep{ferrando2025dynamically} gate or scale the intervention based on the current context, intending to leave benign generation largely untouched; related approaches use probes or lightweight controllers to trigger, scale, or calibrate the intervention \citep{li-etal-2025-fairsteer, wang2025adaptive, cheng2025steering,hegazy-etal-2026-guiding}.
 

However, whether conditional or not, the intervention itself remains dense in space: whenever steering is applied, the full vector is added to all $D$ neurons of the hidden state. We posit that this is more than the task requires, for two reasons. First, many neurons carry little or no information about the concept, so shifting them cannot suppress the undesired behavior and only perturbs the representation. Second, a neuron that does encode the concept may already be in the desired regime at the current step; pushing a non-toxic activation further in the non-toxic direction does not make the output less toxic, but it does move the representation away from the model's familiar activation distribution.

\begin{figure*}[t]
  \centering
  \includegraphics[width=0.95\textwidth]{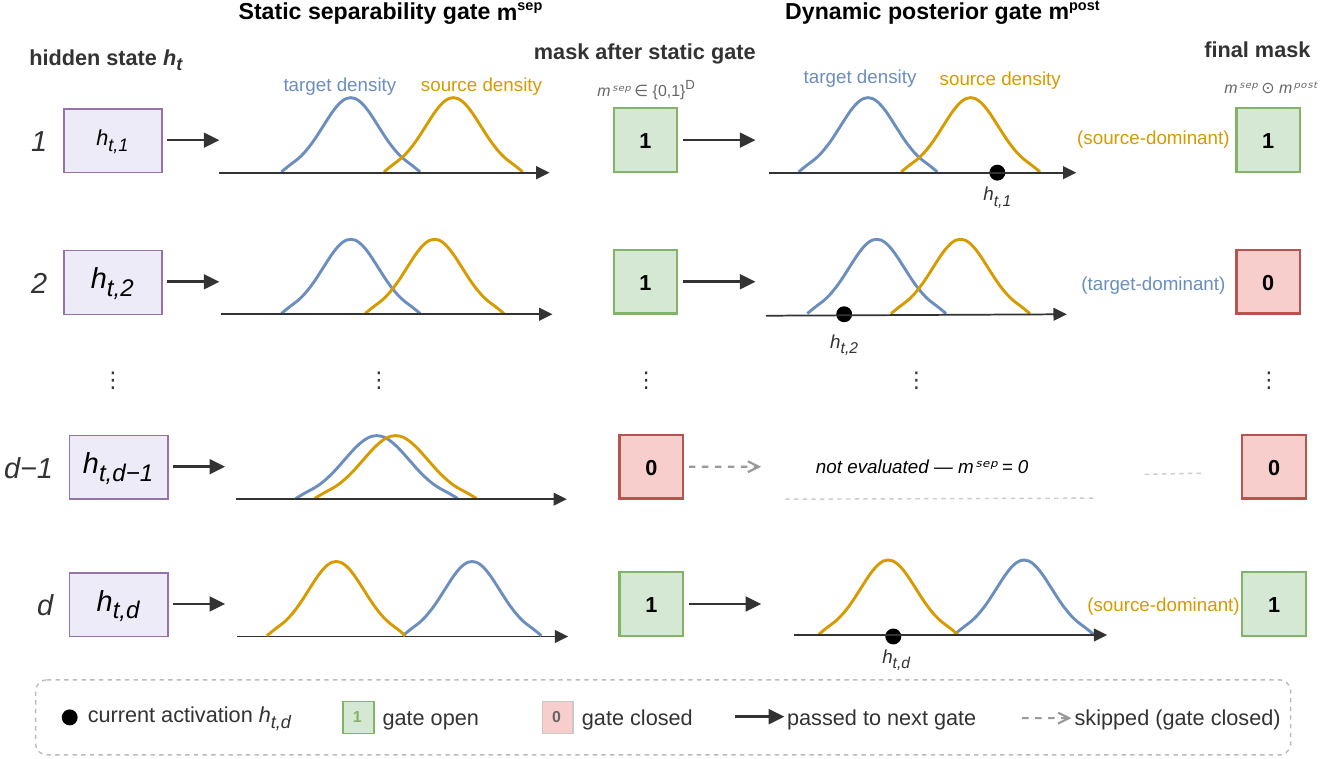}
  \caption{\textbf{The two dimension-level gates of GAPS.} The static separability gate $\mathbf{m}^{\mathrm{sep}}$ identifies which neurons encode the concept, keeping only neurons that separate the target and source concepts (computed offline). The dynamic posterior gate $\mathbf{m}^{\mathrm{post}}$ detects whether the source (undesired) concept is \emph{currently} expressed, steering a neuron only when its current activation is more likely under the source than the target concept. Their product gives the final per-neuron mask applied to the steering vector.}
  \label{fig:method}
  \vspace{-10pt}
\end{figure*}

 
In this work, we add a second axis of conditioning. Token-level methods
decide when to intervene; we also dynamically decide which neurons in the hidden state to intervene on, with the goal of changing the hidden state no more than the
steering objective requires. We do this with two dimension-level gates (Figure~\ref{fig:method}). The
first, a static \emph{separability gate}, restricts steering to only the neurons that carry discernible information about the target concept. For each neuron, we measure how well its activations separate the two concepts with the Area under the ROC, and keep the neuron only if separable enough. 
The second, a dynamic \emph{posterior gate}, decides at each generation step whether a neuron is currently expressing the undesired behavior. We model the
concept-conditional activation distribution of each neuron as a Gaussian and steer a neuron only if its current activation is more likely under the
undesired concept than under the desired one. 

The separability gate is computed once from the contrastive data and fixes
the set of neurons that can encode the behavior at all. The posterior
gate is evaluated at every generation step and decides whether those
neurons are currently expressing the undesired behavior. We refer to the combined gates as GAPS (Gated Activation steering via Posterior and Separability). GAPS strictly generalizes prior methods: with both dimension-level gates set to one, it reduces to CAST or DSAS, and with the token-level gate also set to one, it reduces to unconditional steering.
The gates cost $O(D)$ elementwise operations per token, the same order as
the steering vector addition itself, so dimension-level
conditioning adds only a constant-factor overhead to canonical steering, and all required statistics are estimated from the same contrastive activations used to construct the steering vector. 

We evaluate on toxicity mitigation with RealToxicityPrompts \citep{gehman-etal-2020-realtoxicityprompts} and concept removal over seven OneSeC concepts \citep{scarlini-etal-2019-just}, using Gemma-3 (4B) and Qwen-3 (1.7B) with CAST and DSAS as token-level baselines, sweeping the intervention strength to trace the trade-off against Wikipedia perplexity and MMLU \citep{wikidump, hendryckstest2021}. Across all eight combinations of model, token-level method, and task, GAPS matches or improves the Pareto front of its token-level counterpart. Under a fixed capability budget, DSAS+GAPS lowers Gemma-3's toxicity rate from 6.52\% to 0.48\%, versus 3.52\% for DSAS alone. Ablations attribute most of the gain to the posterior gate, with the separability gate a cheap complement: the strongest operating point requires both.

Our contributions are as follows:
\vspace{-10pt}

 \begin{itemize}
    \item We introduce dimension-level conditioning as a second axis of selectivity in activation steering: beyond deciding \emph{when} to steer, we \emph{dynamically} decide \emph{which} neurons to steer.
    \vspace{-5pt}
    \item We propose GAPS, a pair of training-free gates: a static separability gate that restricts steering to neurons with reliable concept information, and a dynamic posterior gate that steers a neuron only when its activation is better explained by the undesired concept. GAPS plugs into existing conditional methods and strictly generalizes them.
    \vspace{-5pt}
    \item Across two models, two token-level methods, and two tasks, GAPS consistently matches or improves the Pareto front of its token-level counterparts; ablations attribute most of the gain to the posterior gate.
\end{itemize}

\vspace{-10pt}
\section{Preliminaries}
\vspace{-5pt}

\stitle{Neuron.}
A neuron refers to a component of a hidden state vector in a transformer layer. We write $h \in \mathbb{R}^D$ for the hidden state, where neuron $d$ corresponds to the $d$-th coordinate, for $d \in \{1,\dots,D\}$. During generation, $h_t$ denotes the hidden state at generation step $t$, and $h_{t,d}$ denotes the activation value of neuron $d$ at that step.

\stitle{Activation steering.}
Given a steering objective (e.g., non-toxic generation), we collect hidden-state activations from target examples ($\mathrm{tar}$), which represent the behavior we steer toward, and source examples ($\mathrm{src}$), which represent the behavior we steer away from. During generation, standard activation steering adds a fixed direction
to the hidden state: $h_t \leftarrow h_t + \alpha\,v$,
where $\alpha > 0$ controls the intervention strength.
We use the standard mean-difference direction \citep{rimsky-etal-2024-steering}:
\[
  v
  =
  \mu^{\mathrm{tar}}
  -
  \mu^{\mathrm{src}},
  \;
  \mu^{c}
  =
  \frac{1}{N_c}
  \sum_{i=1}^{N_c}
  h^{c}_i,
  \;
  c\in\{\mathrm{tar},\mathrm{src}\},
\]
where $h^{c}_i$ is the activation extracted from the $i$-th
example of concept $c$.

\stitle{Token-level conditional steering.} Rather than applying the intervention at every generation step (every token), token-level conditional methods modulate its strength using a scalar gate
$g(h_t)\in[0,1]$:
\begin{equation}
  h_t \leftarrow h_t + \alpha\, g(h_t)\, v
  \label{eq:token-gated}
\end{equation}
We consider two such methods. \textsc{CAST} \citep{lee2025programming} uses a hard binary gate obtained by thresholding the alignment between the current hidden state and an extracted condition direction. \textsc{DSAS} \citep{ferrando2025dynamically}, on the other hand, uses the output of a probe as a continuous gate that scales the intervention according to the detected strength of the undesired behavior.

\vspace{-5pt}
\section{Neuron-Level Conditional Steering}
\vspace{-5pt}

Token-level conditional methods (Eq.~\ref{eq:token-gated}) address a key weakness of unconditional steering: by modulating the intervention with a token-level gate $g(h_t)$, they intervene only when the running context actually exhibits the source (undesired) behavior, leaving benign generation untouched and thereby reducing unnecessary capability loss. Both \textsc{CAST} and \textsc{DSAS} thus determine \emph{when} to steer. However, once the gate is active, they apply the full dense steering vector $v$ to all $D$ neurons, regardless of whether a neuron carries concept-discriminative information at all, or whether its current activation is already consistent with the target (desired) behavior. Steering such neurons cannot reduce the source behavior; it can only cause unnecessary perturbation of the model. Our method adds neuron-level selectivity to address these two limitations: a static separability gate determines which neurons can encode the behavior, and a dynamic posterior gate determines whether intervention on these neurons is warranted \emph{now} (Figure~\ref{fig:method}).

\stitle{Static separability gate.}
For each neuron $d \in \{1,\dots,D\}$, we measure how well its activation distribution separates the two concepts using the area under the ROC curve, $\mathrm{AUROC}_d = \Pr(h^{\mathrm{src}}_d > h^{\mathrm{tar}}_d)$, estimated from the contrastive sets; a value of $\tfrac12$ indicates no concept information.  Computing it via the Mann--Whitney $U$ identity
\citep{mann1947test,hanley1982meaning,mason2002areas} also yields the exact mean and standard deviation of the estimate under the null hypothesis of no concept signal: $\mathrm{AUROC}_d$ has mean $\tfrac12$ and standard deviation
\begin{equation}
  \sigma_0 \;=\;
  \sqrt{\frac{N_{\mathrm{src}} + N_{\mathrm{tar}} + 1}
             {12\,N_{\mathrm{src}} N_{\mathrm{tar}}}},
  \label{eq:null-sd}
\end{equation}

where $N_c$ denotes the number of contrastive examples of concept $c \in \{\mathrm{src}, \mathrm{tar}\}$.

We retain neuron $d$ only if its deviation from chance exceeds a
two-sided threshold of $\tau_z$ null standard deviations,
\begin{equation}
  \msep_d \;=\;
  \ind{\bigl|\mathrm{AUROC}_d - \tfrac12\bigr| > \tau_z \sigma_0},
  \label{eq:sep-mask}
\end{equation}

where $\tau_z$ controls the stringency of the gate. Unlike a fixed AUROC cutoff, whose statistical meaning depends on sample size, this criterion adapts to sample size: with fewer examples, a neuron must have an AUROC farther from $0.5$ to be retained, whereas with more examples, smaller but statistically reliable deviations suffice. We set $\tau_z = 7$, well above the $z \approx 5.1$ required for Bonferroni-corrected significance at $\alpha_0 = 10^{-3}$ across all $D$ neurons, so the gate retains only neurons whose concept signal is statistically unambiguous (more details in Appendix \ref{app:sensitivity}). With $N_{\mathrm{src}} = N_{\mathrm{tar}} = 700$ (matching our OneSeC setup), this corresponds to retaining only neurons whose AUROC lies outside approximately $[0.39, 0.61]$. The test is two-sided because a neuron may encode the source concept through either larger or smaller activations; the appropriate steering direction is captured by $v$.

\stitle{Dynamic posterior gate.}
The static gate identifies which neurons ever encode the concept; it does not tell us whether a given neuron requires intervention \emph{now}.
Let $C = \{c^{tar}, c^{src}\}$ denote the target and source concepts.
Our goal is to steer neuron $d$ only when its current activation value $h_{t,d}$ is more likely under the source concept $c^{\mathrm{src}}$ than under the target concept $c^{\mathrm{tar}}$.
Using Bayes' theorem, we compute the posterior probability that a given activation $h_{t,d}$ arose from $c^{src}$, under the assumption that both concepts are a priori equally likely:

\vspace{-10pt}

\begin{equation}
\resizebox{0.89\columnwidth}{!}{$
\begin{aligned}
p\bigl(c^{\mathrm{src}} \mid h_{t,d}\bigr)
&=
\frac{
  p\bigl(h_{t,d}\mid c^{\mathrm{src}}\bigr)
  p\bigl(c^{\mathrm{src}}\bigr)
}{
  \displaystyle
  \sum_{c\in C}
  p\bigl(h_{t,d}\mid c\bigr)p(c)
}
\\[3pt]
&\overset{\text{equal priors}}{=}
\frac{
  p\bigl(h_{t,d}\mid c^{\mathrm{src}}\bigr)
}{
  p\bigl(h_{t,d}\mid c^{\mathrm{src}}\bigr)
  +
  p\bigl(h_{t,d}\mid c^{\mathrm{tar}}\bigr)
}.
\end{aligned}
$}
\label{eq:posterior}
\end{equation}

Directly estimating the two concept-conditional activation densities, $p(h_{t,d}\mid c^{\mathrm{src}})$ and $p(h_{t,d}\mid c^{\mathrm{tar}})$, using non-parametric methods (e.g., kernel density or histogram estimators) is unreliable given the limited number of contrastive activation samples per neuron and would incur additional per-token cost during generation. Therefore, following the finding in \citep{haider2026neurons, fereidouni-etal-2026-evaluating} that concept-related neuron activations follow approximately Gaussian distributions, we adopt a parametric Gaussian model for each concept-conditional density:

\vspace{-20pt}

\begin{equation}
  p\bigl(h_{t,d} \mid c\bigr)
  \;=\;
  \mathcal{N}\!\bigl(
    h_{t,d};
    \mu_d^{c},
    (\sigma_d^{c})^2
  \bigr),
  \quad c \in C,
  \label{eq:gauss}
\end{equation}

where $\mu_d^{c} = [\mu^{c}]_d$ is the mean activation of neuron $d$ for concept $c$, and $(\sigma_d^{c})^2$ is the corresponding variance, both estimated from the contrastive samples belonging to concept $c$. 


The posterior gate then steers a neuron only when its current activation is better explained by the source concept:

\vspace{-20pt}

\begin{equation}
  \mctx_{d}(h_t)
  =
  \ind{
    p\bigl(c^{\mathrm{src}}\mid h_{t,d}\bigr)
    >
    \tfrac{1}{2}
  }.
  \label{eq:ctx-gate}
\end{equation}

Intuitively, if the current activation $h_{t,d}$ is more likely under the target-concept distribution than under the source-concept distribution, pushing it further along $v_d$ is unlikely to further suppress the source behavior and may instead introduce an unnecessary perturbation.

\begin{figure*}[t]
  \centering
  \includegraphics[width=\textwidth]{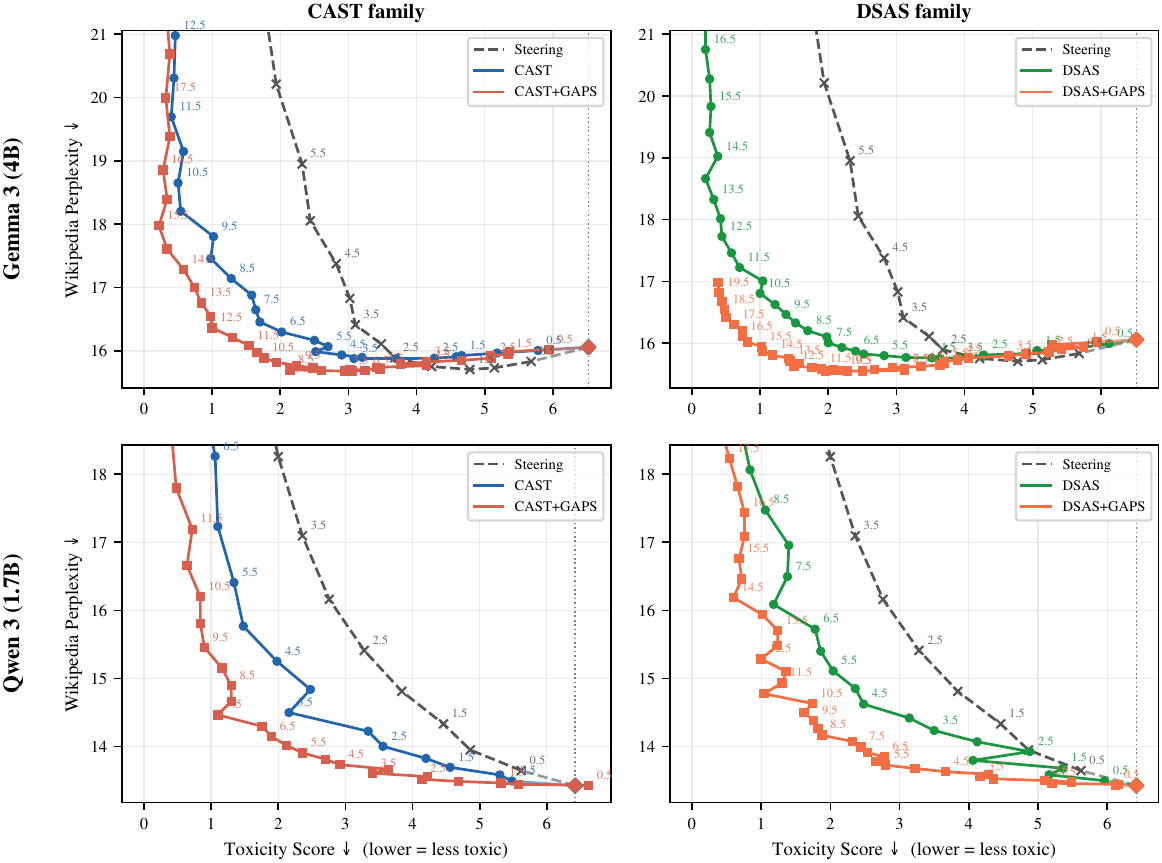}
  \caption{Toxicity mitigation trade-offs: Wikipedia perplexity vs. toxicity score (lower is better on both axes) for Gemma-3 4B (top) and Qwen-3 1.7B (bottom), swept over steering strength $\alpha$. Each panel shows unconditional steering (dashed), a token-level baseline (CAST, left; DSAS, right), and the baseline with our gates (+GAPS); the diamond marks the unsteered model. +GAPS consistently shifts the Pareto front toward the bottom-left.}
  \label{fig:toxicity_ppl}
  \vspace{-10pt}
\end{figure*}

\begin{figure*}[t]
  \centering
  \includegraphics[width=\textwidth]{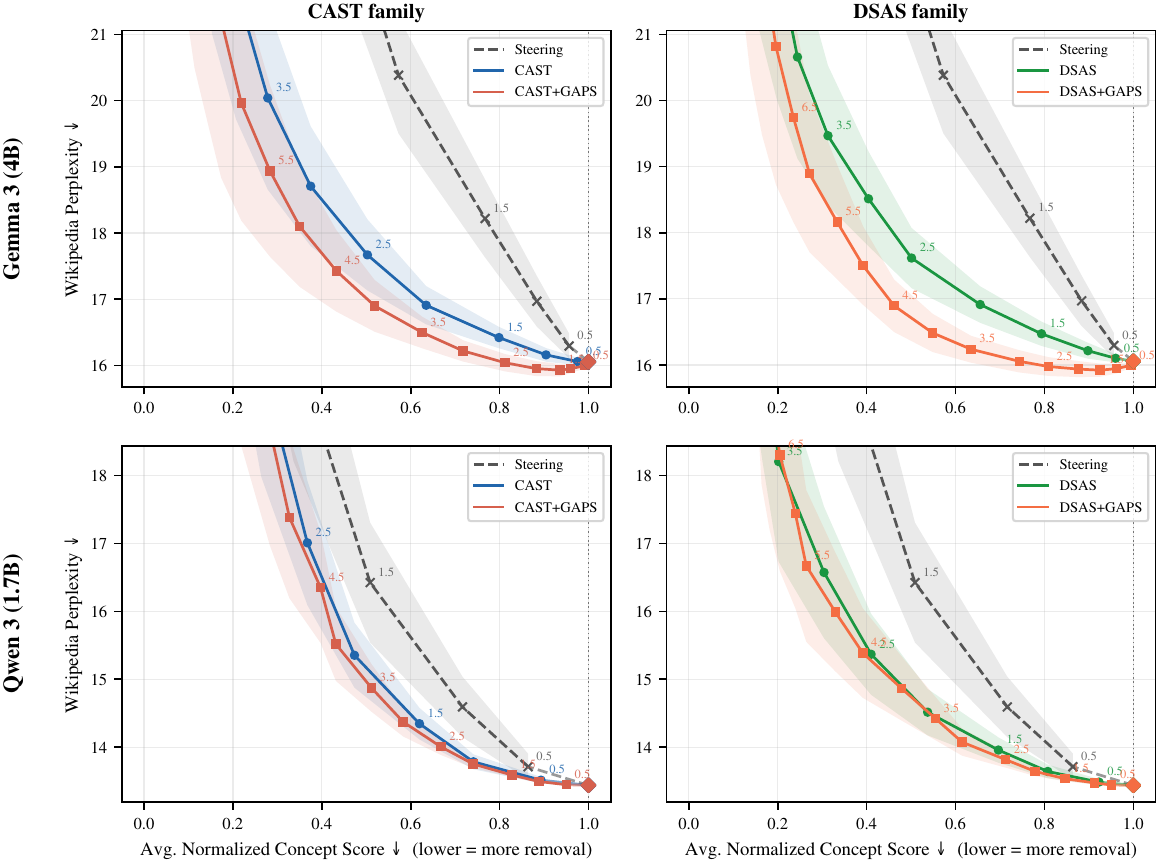}
  \caption{Concept removal trade-offs: Wikipedia perplexity vs.\ average normalized concept score (per-concept score divided by its unsteered value, averaged over seven OneSeC concepts; lower is better on both axes), swept over $\alpha$. Panels as in Figure~\ref{fig:toxicity_ppl}; shaded bands = $\pm$1 standard error of the mean across concepts. +GAPS again matches or improves the trade-off on both models.}
  \label{fig:concept_ppl}
  \vspace{-10pt}
\end{figure*}

\stitle{Full update.}
Combining the token-level gate with both neuron-level conditions
yields intervention:
\begin{equation}
\resizebox{0.89\columnwidth}{!}{$
  \boxed{\;
  h_t \;\leftarrow\; h_t \;+\;
  \alpha\; g(h_t)\;
  \big(\msep \odot \mctx(h_t)\big) \odot v
  \;}
$}
  \label{eq:full}
\end{equation}
where $\odot$ is the elementwise product, $g \in \{g_{\textsc{cast}}, g_{\textsc{dsas}}\}$. We call the combined gate ($\msep\!\odot\!\mctx$), GAPS (yielding CAST+GAPS and DSAS+GAPS). During autoregressive generation, the update is applied to the final token position of the running context. 
Setting $\msep = \mctx = 1$ recovers token-level conditional steering exactly (CAST or DSAS), and additionally setting $g \equiv 1$ recovers unconditional steering, so Eq.~\ref{eq:full} strictly generalizes the prior methods; the sweep over $\alpha$ traces the behavior-vs.-capability Pareto curve for every variant.
Offline, all required statistics are estimated in a single pass over the same contrastive activations used to construct $v$; online, evaluating both gates adds $O(D)$ elementwise operations per token, the same order as the steering-vector addition $\alpha v$ itself (see Appendix~\ref{app:posterior-gate} for details).

It is important to note that the two dimension-level gates are complementary rather than redundant: $\msep$ is a global statement about which neurons encode the behavior at all, while $\mctx$ is a local statement about whether intervention is currently warranted on those neurons.



\section{Experiments and Results}

\begin{table*}[t]
\centering
\small
\setlength{\tabcolsep}{3.5pt}
\resizebox{\linewidth}{!}{%
\begin{tabular}{l ccc ccc ccc ccc}
\toprule
 & \multicolumn{6}{c}{\textit{Toxicity} (\% toxic generations)}
 & \multicolumn{6}{c}{\textit{Concept removal} (\% of unsteered score)} \\
\cmidrule(lr){2-7} \cmidrule(lr){8-13}
 & \multicolumn{3}{c}{Gemma-3 (4B)} & \multicolumn{3}{c}{Qwen-3 (1.7B)}
 & \multicolumn{3}{c}{Gemma-3 (4B)} & \multicolumn{3}{c}{Qwen-3 (1.7B)} \\
\cmidrule(lr){2-4} \cmidrule(lr){5-7} \cmidrule(lr){8-10} \cmidrule(lr){11-13}
Method & Score\,$\downarrow$ & MMLU & PPL & Score\,$\downarrow$ & MMLU & PPL
       & Score\,$\downarrow$ & MMLU & PPL & Score\,$\downarrow$ & MMLU & PPL \\
\midrule
No intervention & 6.52 & 0.586 & 16.06 & 6.42 & 0.539 & 13.43
                & 100.00 & 0.586 & 16.06 & 100.00 & 0.539 & 13.43 \\
\midrule
CAST & 3.70 & 0.570 & 15.88 & 3.56 & 0.550 & 14.00
     & 64.66 & 0.583 & 16.70 & 66.78 & 0.533 & 13.78 \\
\;$+\,\mathbf{m}^{\text{sep}}$ & \underline{2.78} & 0.570 & 16.14 & 3.20 & 0.546 & 13.99
     & 69.88 & 0.583 & 16.61 & 62.36 & 0.538 & 13.94 \\
\;$+\,\mathbf{m}^{\text{rng}}$ & 4.52 & 0.580 & 16.00 & 2.52 & 0.550 & 14.03
     & 61.82 & 0.584 & 16.65 & 70.55 & 0.534 & 13.73 \\
\;$+\,\mathbf{m}^{\text{sep}}\!\odot\mathbf{m}^{\text{rng}}$
     & 4.38 & 0.571 & 16.07 & 2.86 & 0.550 & 13.97
     & 63.88 & 0.584 & 16.66 & 66.10 & 0.537 & 13.96 \\
\;$+\,\mathbf{m}^{\text{post}}$ & 3.06 & 0.571 & 15.47 & \underline{2.16} & 0.542 & 13.97
     & \textbf{46.36} & 0.581 & 16.63 & \underline{61.83} & 0.536 & 13.92 \\
\;$+\,\mathbf{m}^{\text{sep}}\!\odot\mathbf{m}^{\text{post}}$ {\scriptsize(GAPS)}
     & \textbf{2.14} & 0.569 & 15.70 & \textbf{2.12} & 0.546 & 14.02
     & \underline{49.45} & 0.582 & 16.63 & \textbf{61.35} & 0.538 & 13.93 \\
\midrule
DSAS & 3.52 & 0.573 & 15.76 & 4.06 & 0.545 & 13.80
     & 67.96 & 0.581 & 16.63 & 66.38 & 0.535 & 13.82 \\
\;$+\,\mathbf{m}^{\text{sep}}$ & 2.06 & 0.571 & 16.34 & 3.18 & 0.544 & 14.01
     & 71.28 & 0.585 & 16.57 & 63.94 & 0.538 & 13.92 \\
\;$+\,\mathbf{m}^{\text{rng}}$ & 3.80 & 0.573 & 15.95 & 2.60 & 0.547 & 14.09
     & 75.76 & 0.579 & 16.48 & 67.20 & 0.534 & 13.84 \\
\;$+\,\mathbf{m}^{\text{sep}}\!\odot\mathbf{m}^{\text{rng}}$
     & 2.40 & 0.569 & 16.40 & 2.76 & 0.547 & 14.09
     & 72.66 & 0.580 & 16.59 & 60.76 & 0.539 & 13.99 \\
\;$+\,\mathbf{m}^{\text{post}}$ & \underline{1.26} & 0.572 & 15.29 & \textbf{1.86} & 0.540 & 14.08
     & \underline{56.20} & 0.577 & 16.11 & \underline{60.36} & 0.538 & 13.88 \\
\;$+\,\mathbf{m}^{\text{sep}}\!\odot\mathbf{m}^{\text{post}}$ {\scriptsize(GAPS)}
     & \textbf{0.48} & 0.569 & 16.54 & \underline{2.32} & 0.538 & 14.07
     & \textbf{53.05} & 0.574 & 16.25 & \textbf{59.24} & 0.539 & 13.98 \\
\bottomrule
\end{tabular}
}
\caption{Ablation of the separability gate ($\msep$), posterior gate ($\mctx$), and a simpler range gate ($\mathbf{m}^{\text{rng}}$) on toxicity mitigation (left) and OneSeC concept removal (right, averaged over 7 concepts). Per method, we report the lowest behavior score over the $\alpha$ sweep whose capability cost stays within a fixed budget (perplexity $\leq +5\%$, MMLU $\geq -3\%$ vs.\ unsteered). Best score in \textbf{bold}; second-best \underline{underlined}.}
\label{tab:ablation}
\vspace{-5pt}
\end{table*}

\subsection{Experimental Setup}

\stitle{Tasks and models.} We evaluate on toxicity mitigation
(RealToxicityPrompts; \citealp{gehman-etal-2020-realtoxicityprompts})
and concept removal (seven OneSeC concepts;
\citealp{scarlini-etal-2019-just}), using Gemma-3
4B~\cite{gemmateam2025gemma3technicalreport} and Qwen-3
1.7B~\cite{yang2025qwen3}. All interventions are applied at layer 15 of both models; Appendix \ref{app:layer} reports results at layer 18. 
Full setup details are in Appendix \ref{app:setup}.

\stitle{Capability metrics.} To assess whether our method preserves the general capabilities of the language model while avoiding unnecessary interventions, we evaluate model performance along two axes: (1) perplexity on Wikipedia sentences \cite{wikidump}, which measures fluency and language modeling quality, and (2) performance on the Massive Multitask Language Understanding (MMLU) benchmark \cite{hendryckstest2021}, which measures knowledge and reasoning ability. Rather than discrete MMLU accuracy, we report the mean probability assigned to the correct option, which provides a smoother and more sensitive measure of capability degradation under intervention. Both metrics are computed for the unsteered model and after applying each steering method across a range of intervention strengths $\alpha$, allowing us to trace the full behavior vs capability trade-off.

\stitle{Baselines.} As an additional baseline for dimension-level conditioning, we compare against a range gate $m^{\mathrm{rng}}$ \cite{haider2026neurons}, which steers neuron $d$ only when its current activation $h_{t,d}$ lies within the typical activation range of the source concept:
\vspace{-5pt}
\begin{equation}
  \mrng_d(h_t)
  =
  \ind{
    \left|h_{t,d}-\mu_d^{\mathrm{src}}\right|
    \le
    2.5\,\sigma_d^{\mathrm{src}}
  }.
  \label{eq:rng}
\end{equation}

\subsection{Toxicity Mitigation}
\label{sec:toxicity}

Large language models are prone to generating toxic continuations when conditioned on adversarial or toxicity-eliciting prompts \cite{gehman-etal-2020-realtoxicityprompts}. Our goal in this setting is selective intervention: the model should be steered away from toxic generations only when the running context actually carries toxic content, while generation on benign contexts should remain untouched.

We elicit toxic continuations using prompts from the RealToxicityPrompts (RTP) dataset \cite{gehman-etal-2020-realtoxicityprompts}.
We sample $k=10$ continuations per prompt and classify each as toxic or non-toxic using a RoBERTa-based toxicity classifier. 
This protocol follows common practice in the literature \citep{rodriguez2025controlling, ferrando2025dynamically, suau2024whispering, rodriguez2025lineas}.
We report the toxicity rate: the fraction of continuations labeled toxic, averaged over prompts and expressed as a percentage. The protocol is applied to the unsteered model and to every steering configuration, ensuring that toxicity reductions are directly comparable across all setups.

As shown in Figure~\ref{fig:toxicity_ppl}, both token-level conditional steering baselines improve substantially over unconditional steering. On both models, DSAS attains lower perplexity than unconditional steering at matched toxicity levels, and CAST exhibits the same pattern, tracing a better Pareto front. This confirms that deciding \emph{when} to steer already reduces unnecessary capability loss. Our approach adds a complementary axis of selectivity: it conditions the intervention at the dimension level via the separability gate and posterior gate ($\msep \odot \mctx$), deciding which neurons to steer. 
Applying our gates on top of either baseline (CAST+GAPS, DSAS+GAPS) consistently shifts the Pareto front further toward the bottom-left: at any given toxicity score, the gated variants result in equal or lower perplexity than their token-level counterparts, across both Gemma-3 (4B) and Qwen-3 (1.7B) and both token-level methods. In addition to perplexity, we report MMLU against the toxicity score in Figure~\ref{fig:toxicity_mmlu} (Appendix~\ref{app:mmlu_toxicity}). The benefit of our gates is most pronounced for Gemma-3 (4B), where the gated variants retain substantially higher MMLU than token-level and unconditional steering at matched toxicity levels, degrading more gracefully as the intervention strength increases; on Qwen-3 (1.7B), all methods perform comparably on MMLU across most of the toxicity range.

\subsection{Concept Removal}
\vspace{-5pt}

Beyond toxicity mitigation, we evaluate our method on concept removal. As in the toxicity setting, the goal is selective intervention: steer away from a source concept (e.g., \emph{baby}) only when the running context evokes it, leaving unrelated contexts unaffected.
For the contrastive data, we follow \citet{rodriguez2025controlling} and use the OneSeC dataset~\cite{scarlini-etal-2019-just} with the same seven concepts (\emph{football}, \emph{cloud}, \emph{baby}, \emph{church}, \emph{book}, \emph{flower}, \emph{balloon}) and sampling protocol: for each concept, 700 samples of the source concept (to steer away from) and 700 samples drawn from the remaining concepts (to steer toward).
To measure removal, we use GPT-4o-mini as an LLM judge to determine whether the source concept is present in each generated continuation; the prompt is in Appendix~\ref{app:judge_prompt}.

Figure~\ref{fig:concept_ppl} reports the perplexity--removal trade-off. As in the toxicity setting, applying dimension-level gates on top of either token-level baseline matches or improves the Pareto front on both models, with the largest gains on Gemma-3 (4B).

MMLU results are reported in Figure~\ref{fig:onesec_mmlu} (Appendix~\ref{app:mmlu_onesec}). The two capability metrics paint complementary pictures. On Gemma-3 (4B), where our gates yield the largest perplexity gains, the dimension-gated variants perform nearly identically to their token-level counterparts on MMLU. On Qwen-3 (1.7B), the pattern reverses: although the perplexity gains were modest, the gated variants retain higher MMLU at matched removal levels for most intervention strengths. Overall, dimension-level gating improves the trade-off along at least one capability axis on both models; more discussion in Appendix~\ref{app:mmlu_onesec}.

\subsection{Analysis of Dimension-Level Gates}
\label{sec:ablation}








 
Table~\ref{tab:ablation} isolates the contribution of each dimension-level condition. For every method, we sweep the steering strength $\alpha$ and report the operating point with the lowest source (undesired) concept score whose capability cost stays inside a fixed budget: perplexity at most 5\% above, and MMLU correct-option probability at most 3\% below, the unsteered model. All rows are therefore compared at matched capability cost.
For concept removal, the capability budget is defined once from the unsteered model, and for each of the seven OneSeC concepts, we separately select the steering strength with the lowest concept score within the budget. We normalize the resulting concept score by its unsteered value, so that the unsteered baseline is 100\%, and then average the normalized scores across all seven concepts. MMLU and perplexity are likewise averaged over the per-concept operating points.

This gives $8$ independent settings: $2$ models $\times$ $2$ token-level methods (CAST, DSAS) $\times$ $2$ tasks. Throughout, $\msep$ is \emph{static}, computed once from the contrastive sets, whereas $\mrng$ and $\mctx$ are both \emph{dynamic}: each is evaluated at the current activation $h_{t,d}$ at every generation step, and the two differ only in their decision rule, membership in the undesired concept's empirical range (Eq.~\ref{eq:rng}) versus the posterior $p\bigl(c^{\mathrm{src}}\mid h_{t,d}\bigr) > \tfrac{1}{2}$ (Eq.~\ref{eq:ctx-gate}). We summarize four findings.

\stitle{Overall ranking: posterior-gated variants perform best.} In all $8$ settings, the best-performing variant contains the posterior gate, and in $7$ of $8$, the two posterior-gated variants occupy \emph{both} the best and second-best positions ($15$ of the $16$ top-two slots). The single exception is Gemma-3 with CAST on toxicity, where $\msep$ alone ($2.78$) edges out $\mctx$ ($3.06$); $\mboth$ is still the best variant there ($2.14$). The strongest overall result is obtained by DSAS$+$GAPS on Gemma-3, which lowers toxicity from $6.52\%$ (unsteered) to $0.48\%$. By comparison, DSAS alone reduces toxicity only to $3.52\%$. Thus, GAPS achieves a $92.6\%$ reduction relative to the unsteered model, versus $46.0\%$ for the token-level gate alone.

\stitle{Posterior vs.\ separability: knowing where the concept lives is not the same as knowing where to act.} 
The posterior gate ($\mctx$) outperforms the static separability gate ($\msep$) in $7$ of the $8$ settings, with improvements of up to a $41\%$ relative reduction in score (toxicity $3.18\%\!\rightarrow\!1.86\%$ for Qwen-3 with DSAS). The two methods answer different questions. The static gate ($\msep$) identifies dimensions that \emph{can} encode the undesired concept and steers all of them whenever the token-level gate is active. The posterior gate ($\mctx$) instead identifies dimensions that are \emph{currently} expressing the undesired concept by comparing how well the present activation is explained by the undesired and desired concept distributions. Its consistent advantage shows that knowing which neurons can encode a behavior is insufficient; effective steering also requires knowing when that encoding is actually active.

\stitle{Posterior vs.\ range: why range falls short.} 
The posterior gate consistently outperforms the range gate. Across all 8 settings, $\mctx$ achieves a lower score than $\mrng$, and combining it with the separability gate ($\mboth$, i.e.\ GAPS) likewise outperforms $\msep\!\odot\!\mrng$ in every case. The difference lies in the decision rule. The range gate ($\mrng$) considers only the undesired-concept distribution, activating whenever $h_{t,d}$ falls within $2.5$ standard deviations of the undesired-concept mean (Eq.~\ref{eq:rng}). As a result, it also steers activations in regions where the desired and undesired distributions overlap, even when the current activation is equally or better explained by the desired concept. The posterior gate instead compares the two concept-conditional densities directly and intervenes only when the undesired concept is the more likely explanation, avoiding these unnecessary updates.

\stitle{The separability gate is a consistently useful, essentially free complement.}
The static separability gate improves whichever decision rule it is paired with. Applied directly on top of the token-level baselines, $\msep$ lowers the score in $6$ of $8$ settings; combined with the range gate, $\msep\!\odot\!\mrng$ improves over $\mrng$ in $5$ of $8$; and combined with the posterior gate, $\msep\!\odot\!\mctx$ (GAPS) improves over $\mctx$ in $6$ of $8$. Notably, the strongest operating point in Table~\ref{tab:ablation} (toxicity reduced from $6.52\%$ to $0.48\%$ with DSAS on Gemma-3) is reached only with the separability gate: $\mctx$ alone plateaus at $1.26\%$. Since $\msep$ is computed once from the contrastive sets and adds no per-token cost, it offers these gains at effectively zero overhead.

\section{Related Works}

\stitle{Activation steering.} Activation steering adds a concept direction to the hidden state during inference, with no gradient updates and little inference cost. The direction is typically extracted from contrastive examples or probes \citep{turner2025steering,rimsky-etal-2024-steering,zou2023transparency,li2023inferencetime}, while later work optimizes the intervention itself \citep{rodriguez2025controlling,rodriguez2025lineas}. All of these methods are unconditional and dense: the intervention is applied at every generation step and to all dimensions of the hidden state.

\stitle{When to steer.} A first axis of selectivity decides \emph{when} to intervene, per input or per generation step. CAST \citep{lee2025programming} applies a hard gate based on the alignment between the current hidden state and an extracted condition direction, and DSAS \citep{ferrando2025dynamically} scales the intervention with the output of a trained probe. Several works follow the same pattern, using probes or controllers to trigger, scale, or calibrate the intervention
\citep{li-etal-2025-fairsteer,wang2025adaptive,cheng2025steering,hegazy-etal-2026-guiding, hedstrom2025to}. In all cases, however, the intervention remains dense: the steering vector is applied in full at the chosen site, with no dimension-level selectivity.








\stitle{Static dimension selection.} Some methods restrict the intervention to a subset of dimensions, but this selection is made offline: AurA \citep{suau2024whispering} dampens experts chosen by AUROC, and SADI \citep{wang2025semanticsadaptive} applies a top-$K$ contrastive mask.
Neither decides online, per generation step and per neuron, whether steering is currently warranted, the axis of selectivity GAPS introduces.

\vspace{-5pt}
\section{Conclusion}
\vspace{-5pt}

We introduced dimension-level conditioning as a second axis of selectivity in activation steering: beyond deciding when to intervene, our gates decide, at each generation step, which neurons in the hidden state to steer. The static separability gate restricts steering to neurons carrying reliable concept information, and the dynamic posterior gate intervenes only when a neuron's current activation is better explained by the undesired concept. The combined gates, GAPS, are training-free, add negligible overhead, and strictly generalize existing conditional methods. Across two models, two token-level methods, and two tasks, GAPS matches or improves the Pareto front, reducing Gemma-3's toxicity from 6.52\% to 0.48\% at matched capability cost. Ablations attribute most of the gain to the posterior gate: knowing which neurons can encode a concept is not enough; effective steering requires knowing when it is active.


\section*{Limitations}

The dynamic posterior gate assumes equal priors over the source and target concepts (Eq.~\ref{eq:posterior}).
In deployment, a practitioner who expects mostly benign contexts could set $p(c^{\mathrm{src}}) < \tfrac{1}{2}$ to make the gate more conservative, whereas adversarial settings may warrant the opposite. 
We leave a systematic study of prior selection to future work.


\bibliography{custom}

\appendix


\begin{table*}[t]
\centering
\small
\begin{tabular}{p{0.13\textwidth}p{0.80\textwidth}}
\toprule
\textbf{Concept} & \textbf{Definition} \\
\midrule
Football & Any of various games played with a ball (round or oval) in which two teams try to kick or carry the ball into each other's goal. \\

Cloud & A visible mass of water or ice particles suspended at a considerable altitude. \\

Baby & A very young child (birth to 1 year) who has not yet begun to walk or talk. \\

Church & One of the groups of Christians who have their own beliefs and forms of worship. \\

Book & A written work or composition that has been published (printed on pages bound together). \\

Flower & Reproductive organ of angiosperm plants especially one having showy or colorful parts. \\

Balloon & Large tough nonrigid bag filled with gas or heated air. \\

\bottomrule
\end{tabular}
\caption{WordNet definitions supplied to the LLM judge for the seven OneSeC concepts.}
\label{tab:concept-definitions}
\end{table*}

\FloatBarrier

\section{Implementation of the Posterior Gate}
\label{app:posterior-gate}
For numerical stability, we evaluate the posterior of Eq.~\ref{eq:posterior}, with the Gaussian densities of Eq.~\ref{eq:gauss}, entirely in log space.


\stitle{Concept-conditional log-densities.} Substituting the Gaussian model of Eq.~\ref{eq:gauss}, i.e.,
$p\bigl(x_d \mid c\bigr) = \bigl(2\pi (\sigma^{c}_{d})^{2}\bigr)^{-1/2}
\exp\!\left(-\frac{(x_d - \mu^{c}_{d})^{2}}{2(\sigma^{c}_{d})^{2}}\right)$,
the log-density of neuron $d$ under concept $c$ at the observed activation $x_d = h_{t,d}$ is

\begin{equation}
\begin{aligned}
  L^{c}_{d}
  \;:=\;
  &\log p\bigl(x_d \mid c\bigr) \\
  =\;&
  -\,\frac{\bigl(x_d - \mu^{c}_{d}\bigr)^{2}}
          {2\,(\sigma^{c}_{d})^{2}}
  \;-\;
  \frac{1}{2}\log\!\bigl(2\pi\,(\sigma^{c}_{d})^{2}\bigr).
\end{aligned}
\label{eq:app-logpdf}
\end{equation}

The normalization term  $\tfrac{1}{2}\log\bigl(2\pi(\sigma^{c}_{d})^{2}\bigr)$ is precomputed offline once per neuron and concept, so the online cost of Eq.~\ref{eq:app-logpdf} is one squared difference and one division per neuron.

\stitle{Log-posterior.} Taking the logarithm of Eq.~\ref{eq:posterior} and substituting Eq.~\ref{eq:app-logpdf}, the log-posterior of the source concept is

\[
  \log p\bigl(c^{\mathrm{src}} \mid x_d\bigr)
  \;=\;
  L^{\mathrm{src}}_{d}
  \;-\;
  \log\!\bigl(e^{L^{\mathrm{src}}_{d}} + e^{L^{\mathrm{tar}}_{d}}\bigr),
  \label{eq:app-logpost}
\]

The denominator is evaluated with a numerically stable log-sum-exp. 

\stitle{Computational cost.}
The posterior gate adds no trained components and no additional forward passes; its entire cost is a fixed number of elementwise operations per neuron. Offline, the per-neuron statistics $\{\mu^{c}_{d}, (\sigma^{c}_{d})^{2}\}$ of Eq.~\ref{eq:gauss} are estimated in a single pass over the same contrastive activations already used to construct the steering vector $\mathbf{v}$, and the memory footprint at generation time is six $D$-dimensional vectors (a mean, a variance, and a precomputed normalization term per concept), i.e.\ $O(D)$. Online, evaluating the gate at a generation step amounts to two evaluations of Eq.~\ref{eq:app-logpdf}, one log-sum-exp, one exponentiation, and one comparison per neuron, a constant number of elementwise operations, computed for all $D$ neurons simultaneously as vectorised tensor operations. The per-token cost is therefore $O(D)$, the same order as the steering-vector addition $\alpha\mathbf{v}$ itself and as the token-level gates it composes with: dimension-level conditioning adds only a constant-factor overhead to conditional steering, with no dependence on the number of contrastive examples at inference time.

\section{Concept Classification}
\label{app:judge_prompt}

To determine whether a generated continuation refers to the source concept, we use an LLM-based binary classifier. For each of the seven OneSeC concepts---\emph{football}, \emph{cloud}, \emph{baby}, \emph{church}, \emph{book}, \emph{flower}, and \emph{balloon}---we provide the classifier with the corresponding WordNet definition and the generated continuation. Supplying the definition allows the classifier to identify semantic references to the concept even when the concept word itself does not appear explicitly.

The classifier is instructed to return only \texttt{yes} or \texttt{no}. We use greedy decoding with temperature~$0$ and allow at most five output tokens. The exact prompt is:

\begin{quote}
\small
\ttfamily
You are a chatbot who answers whether the provided sentence is referring to \{concept\} defined as \{wordnet\_def\} Note that the sentence might not contain the word \{concept\}, but may just be referencing concept as defined. Answer only with 'yes' or 'no'.
\{generated\_text\}
\end{quote}

The WordNet definitions used for the seven concepts are shown in Table~\ref{tab:concept-definitions}. Our prompt is adapted from \citet{rodriguez2025controlling}.

\begin{figure*}[t]
  \centering
  \includegraphics[width=\textwidth]{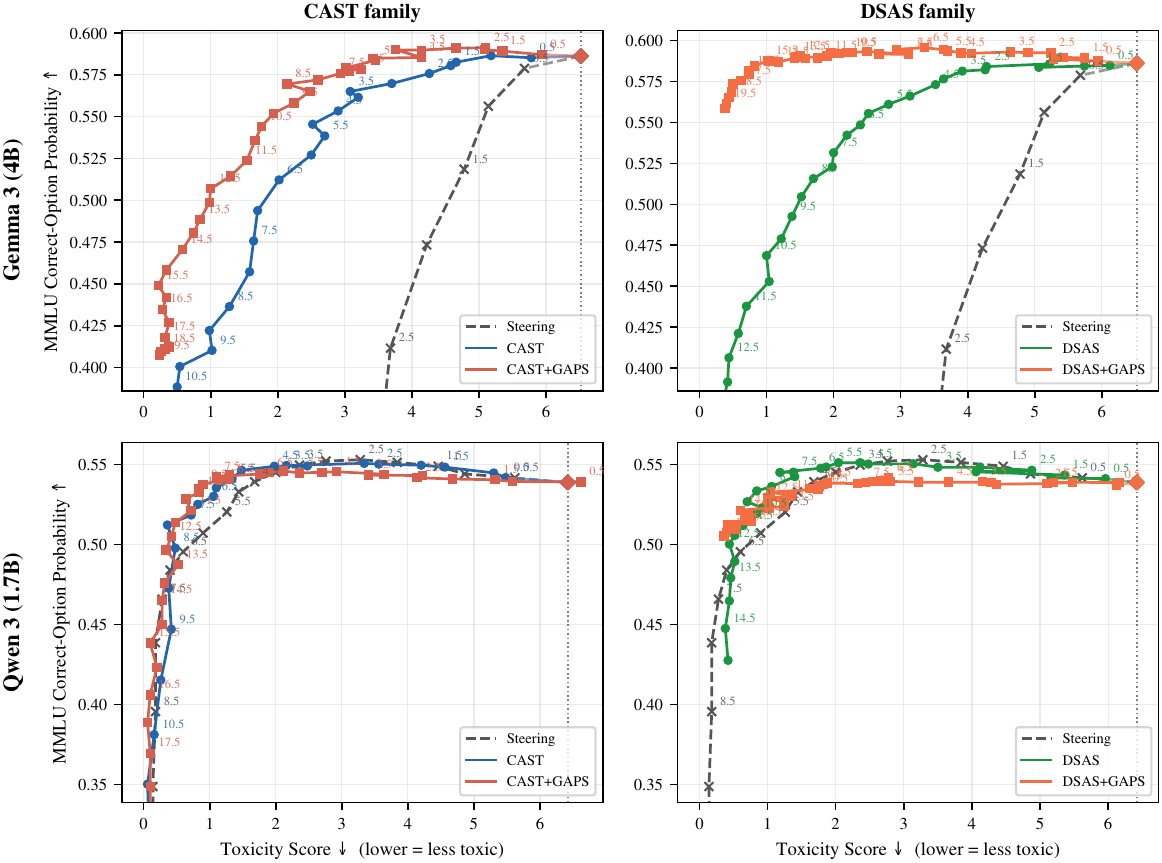}
  \caption{Toxicity mitigation trade-off curves on MMLU. MMLU correct option probability (capability, higher is better) versus toxicity score (lower = less toxic) for Gemma-3 (4B, top) and Qwen-3 (1.7B, bottom), obtained by sweeping the steering strength $\alpha$. Each panel compares unconditional steering (dashed), the token-level conditional baseline (CAST, left; DSAS, right), and the same baseline augmented with our dimension-level gates (+GAPS). Curves closer to the top-left corner are better; the diamond marks the unsteered model. On Gemma-3 (4B), the dimension-gated variants retain higher MMLU performance than their token-level counterparts at matched toxicity levels, degrading more gracefully as the intervention strength increases; on Qwen-3 (1.7B), all methods perform comparably across most of the toxicity range.}
  \label{fig:toxicity_mmlu}
\end{figure*}

\FloatBarrier

\section{MMLU Results - Toxicity}
\label{app:mmlu_toxicity}

Figure~\ref{fig:toxicity_mmlu} reports MMLU correct-option probability against the toxicity score for both models and both steering families.

On Gemma-3 (4B), the dimension-gated variants (+GAPS) retain higher MMLU than their token-level counterparts at matched toxicity levels, with the largest margin in the DSAS. Whereas CAST and DSAS degrade noticeably as the intervention strength $\alpha$ increases, the GAPS variants remain close to the unsteered baseline across most of the $\alpha$ range and degrade more gracefully at the strongest interventions, reaching low toxicity scores at a substantially smaller MMLU cost.

On Qwen-3 (1.7B), all methods perform comparably across most of the toxicity range. At moderate intervention strengths, unconditional steering and the token-level baselines slightly exceed the unsteered MMLU value. At large $\alpha$, however, their performance drops steeply, while the GAPS variants degrade more slowly at high intervention strengths.



\begin{figure*}[t]
  \centering
  \includegraphics[width=\textwidth]{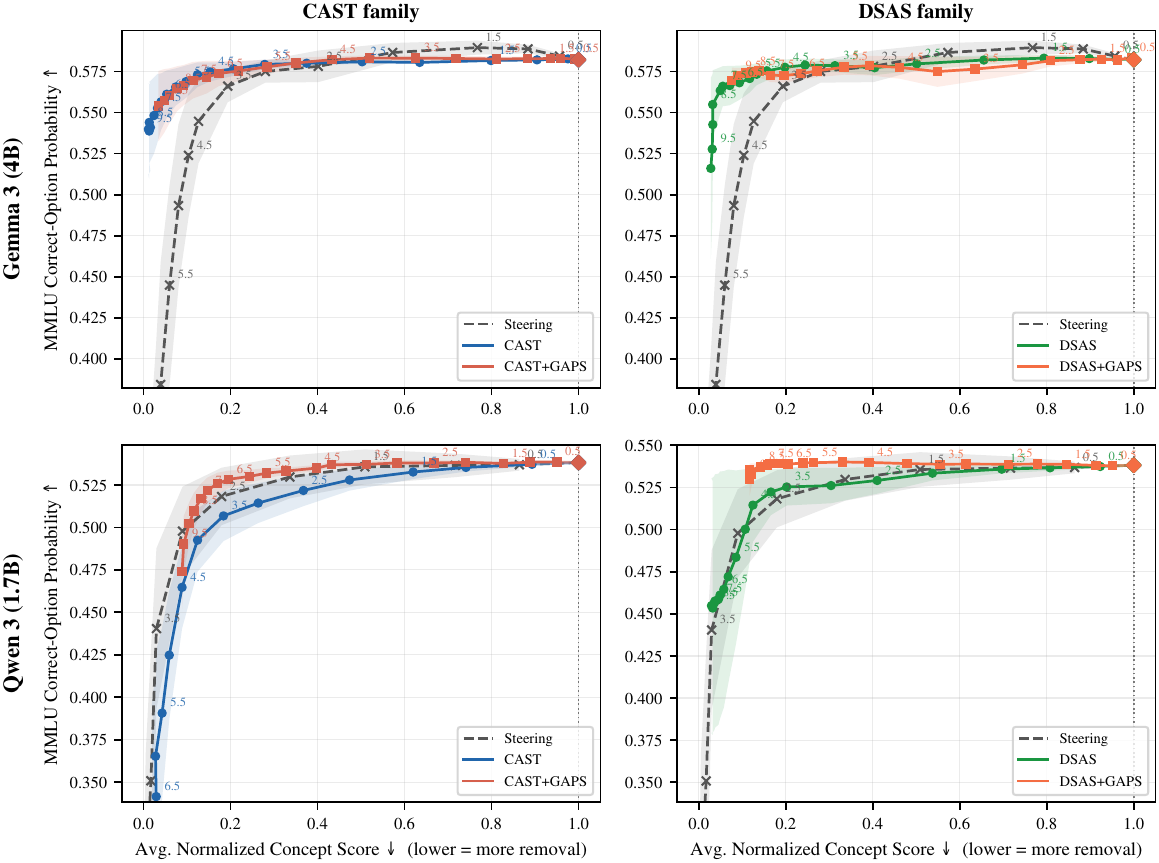}
  \caption{Concept removal trade-off curves on MMLU. MMLU correct-option probability (capability, higher is better) versus average normalized concept score (each concept's score divided by its unsteered value, then averaged over the seven OneSeC concepts; lower = more concept removal), obtained by sweeping the steering strength $\alpha$, for Gemma-3 (4B, top) and Qwen-3 (1.7B, bottom). Each panel compares unconditional steering (dashed), the token-level conditional baseline (CAST, left; DSAS, right), and the same baseline augmented with our dimension-level gates (+GAPS). Curves closer to the top-left corner are better; the diamond marks the unsteered model. Shaded bands show $\pm 1$ standard error of the mean across the seven concepts. On Gemma-3 (4B), the dimension-gated variants perform nearly identically to their token-level counterparts, while on Qwen-3 (1.7B) they retain clearly higher MMLU at matched removal levels for low to moderate intervention strengths.}
  \label{fig:onesec_mmlu}
\end{figure*}

\FloatBarrier

\section{MMLU Results - OneSeC}
\label{app:mmlu_onesec}

Figure~\ref{fig:onesec_mmlu} reports MMLU correct-option probability against the average normalized concept score for both models and both token-level gating methods.

On Gemma-3 (4B), both token-level baselines (CAST and DSAS) improve over unconditional steering, which degrades steeply once the intervention strength $\alpha$ becomes large, while the conditional methods remain near the unsteered MMLU value across most of the removal range. Because the token-level gates already leave little capability on the table, adding our dimension-level gates yields no further gain here: the +GAPS variants perform nearly identically to their token-level counterparts.

On Qwen-3 (1.7B), the picture differs. Token-level gating alone is less effective: CAST, in particular, falls below unconditional steering at strong intervention strengths. Adding the dimension-level gates improves both families, with CAST+GAPS and DSAS+GAPS retaining clearly higher MMLU than their token-level counterparts at matched removal levels for low to moderate intervention strengths. At the strongest interventions, CAST+GAPS also declines slightly; since the token-level gate is applied before our dimension-level gates, the GAPS variants inherit part of the degradation introduced by the underlying CAST gate and cannot fully recover the unconditional-steering performance in this regime.

\begin{figure*}[t!]
  \centering
  \includegraphics[width=\textwidth]{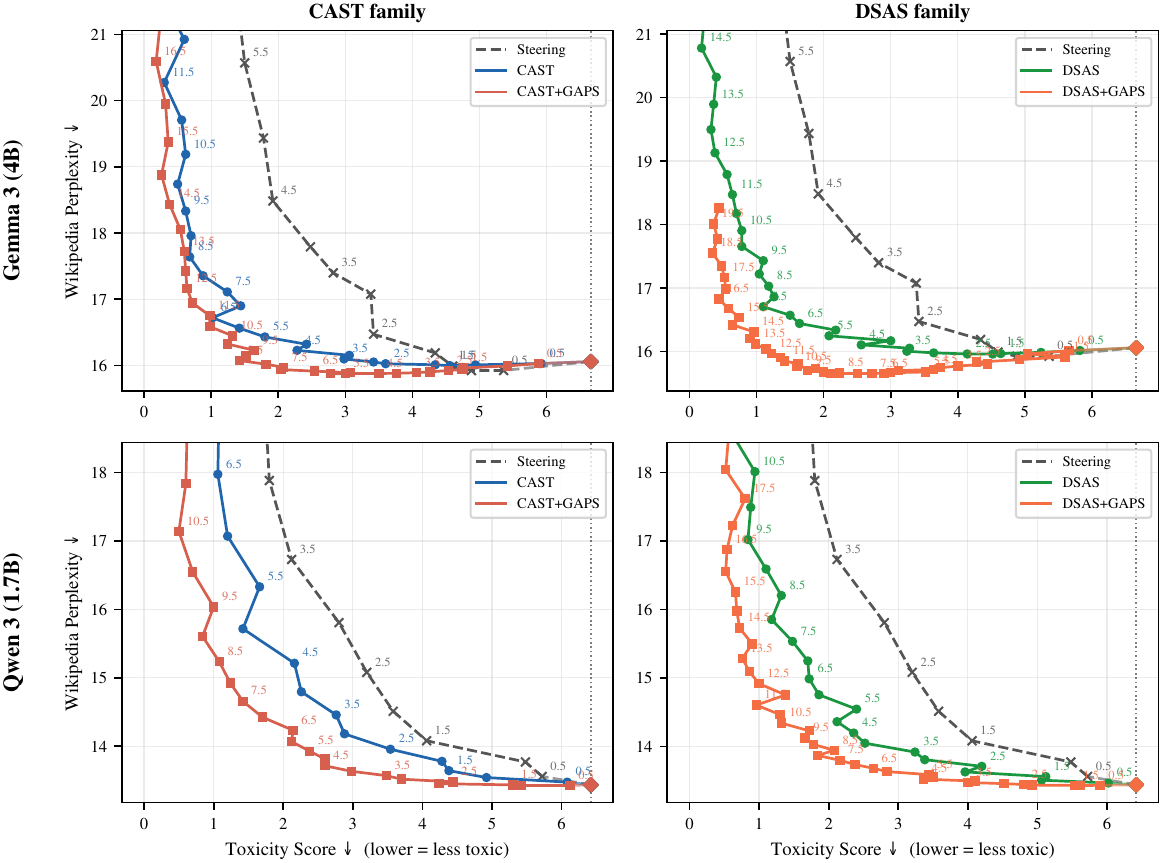}
  \caption{Toxicity mitigation trade-off curves with interventions applied at layer 18. Setup as in Figure \ref{fig:toxicity_ppl}, but steering at layer 18 rather than layer 15. Our dimension-level gates again shift the Pareto frontier toward the bottom-left for both models and both steering families, indicating that the gains are robust to the choice of intervention layer.} 
  \label{fig:layer18}
\end{figure*}

\FloatBarrier

\section{Robustness to the Intervention Layer} 
\label{app:layer}

To verify that the benefits of dimension-level gating are not specific to a single intervention site, we repeat the toxicity mitigation experiment of Section~\ref{sec:toxicity} with all interventions applied at layer~18 instead of layer~15, keeping every other component of the protocol unchanged. (contrastive data, steering vectors, gate statistics, and evaluation setup are all recomputed at the new layer).
Figure~\ref{fig:layer18} reports the resulting perplexity--toxicity trade-off curves for both models and both token-level methods.

The picture closely mirrors the layer-15 results of Figure~\ref{fig:toxicity_ppl}. Token-level conditional steering (CAST, DSAS) again improves substantially over unconditional steering, and adding our dimension-level gates (+GAPS) shifts the Pareto frontier further toward the bottom-left corner: at matched toxicity levels, GAPS attains equal or lower perplexity than their token-level counterparts, across both Gemma-3 (4B) and Qwen-3 (1.7B) and both token-level methods. 

The same qualitative ordering (unconditional $<$ token-level $<$ token-level+GAPS) holds at a different layer, indicating that the gains from dimension-level conditioning are not an artifact of a particular intervention site. 
This is expected under our formulation: both gates are estimated directly from the contrastive activations of whichever layer is steered, so the separability statistics and concept-conditional Gaussians adapt automatically to the chosen layer.
We conclude that the proposed GAPS is robust to the choice of intervention layer.

\section{Implementation Details}
\label{app:setup}
\stitle{Dataset Statistics.} We report the data used to construct the steering directions and evaluate each task. For toxicity steering, we use RealToxicityPrompts~\citep{gehman-etal-2020-realtoxicityprompts}, labeling prompts with toxicity scores $\geq 0.5$ as toxic and those with scores $<0.2$ as non-toxic. After shuffling with a fixed seed, we reserve the first 500 toxic prompts for evaluation, generating 10 completions per prompt, and use the next 5{,}000 examples from each class to collect activations. These subsets are disjoint. For concept steering, following Rodriguez et al.~\citep{rodriguez2025controlling}, we construct contrastive sets for seven OneSeC concepts~\citep{scarlini-etal-2019-just}, using 700 concept-positive and 700 concept-negative sentences per concept to collect activations. We evaluate each concept on 1{,}000 generated completions and measure the proportion that refer to the source concept using the classifier described in Appendix~\ref{app:judge_prompt}. 

\stitle{Model Specifications.} As described in the main text, we use the instruction-tuned Gemma-3 model with 4 billion parameters (Hugging Face name: \texttt{google/gemma-3-4b-it}) and the Qwen-3 model with 1.7 billion parameters (Hugging Face name: \texttt{Qwen/Qwen3-1.7B}). The hidden-state dimensionality, over which our dimension-level gates (GAPS) operate, is $D=2560$ for Gemma-3 (4B) and $D=2048$ for Qwen-3 (1.7B). All interventions are applied to the residual stream (\texttt{hook\_resid\_post}) at layer 15 (layer 18 for the robustness analysis in Appendix \ref{app:layer}).

\stitle{Package Details.}
Our implementation is built on PyTorch \citep{NEURIPS2019_bdbca288}. Models are loaded and hooked through the TransformerLens library \citep{nanda2022transformerlens}, which we use to cache residual-stream activations and to apply all steering interventions during generation, in combination with the HuggingFace Transformers \citep{wolf-etal-2020-transformers} and Datasets \citep{lhoest-etal-2021-datasets} libraries for tokenizers and data loading. Statistical computations use NumPy \citep{harris2020array} and SciPy \citep{virtanen2020scipy}; scikit-learn \citep{pedregosa2011scikit} is used for the PCA and logistic-regression probe of the DSAS baseline.
The LLM-judge evaluation for concept removal (Appendix~\ref{app:judge_prompt}) queries GPT-4o-mini accessed via the OpenRouter API \cite{openrouter}, and all figures are produced with Matplotlib \citep{Hunter2007MatplotlibA2}.

\stitle{Computation Details.}
All experiments were conducted on a high-performance computing (HPC) cluster managed via the Slurm workload manager. We used NVIDIA A40 (48 GB) GPUs for activation collection, steering interventions, and downstream evaluation (toxicity, perplexity, and MMLU). Each job was allocated 1 GPU, 8 CPU cores, and 64 GB of RAM on a single node.

\stitle{Result Reliability.}
Each reported number is an aggregate over a large number of stochastic generations: toxicity scores average over 500 prompts with $k=10$ sampled continuations each (5{,}000 generations per operating point), and concept scores average over 1{,}000 sampled completions per concept before being averaged across concepts. Figures~\ref{fig:concept_ppl} and~\ref{fig:onesec_mmlu} additionally report $\pm 1$ standard error of the mean across the OneSeC concepts. The capability metrics (Wikipedia perplexity and MMLU correct-option probability) are deterministic, given the model, and require no aggregation over runs.

\section{Separability Threshold}
\label{app:sensitivity}

The threshold $\tau_z$ of Eq.~\ref{eq:sep-mask} is the free parameter of the separability mask. Its admissible range is bounded from below by statistical calibration: a Bonferroni-corrected two-sided test at family-wise level $\alpha_0 = 10^{-3}$ across all $D$ dimensions requires $\tau_z \approx 5.0$ ($D = 2048$, Qwen-3) to $5.1$ ($D = 2560$, Gemma-3). 

Any value above this floor, therefore, retains only dimensions with a statistically reliable class signal; where to place $\tau_z$ within the admissible range is an empirical question, which we settle with a sensitivity sweep.

We sweep $\tau_z$ on a single setting, toxicity mitigation with Gemma-3 (4B) and DSAS+GAPS at layer 15, and reuse the selected threshold unchanged in every other experiment (both models, both gating families, both tasks).

For each $\tau_z \in \{5, 6, 7, 8, 9\}$, we sweep the intervention strength $\alpha$ and evaluate toxicity, Wikipedia perplexity, and MMLU exactly as in the main experiments. We then select operating points under two protocols. First, we use the fixed capability budget of the main results: the lowest toxicity whose capability cost stays within perplexity $\leq 1.05\times$ and MMLU $\geq 0.97\times$ the unsteered model. Second, to remove the dependence on any single budget choice, we repeat the selection for $15$ budgets (relative perplexity increase in $\{2, 3, 5, 7.5, 10\}\%$ crossed with relative MMLU drop in $\{1, 3, 5\}\%$) and report the mean and standard deviation of the resulting toxicity rates.

Table~\ref{tab:tauz-sweep} shows that toxicity is minimized in a band around $\tau_z \in [7, 8]$, with degradation on both sides under both protocols. Below the band, at $\tau_z = 6$, fixed-budget toxicity rises to $0.56\%$ and the budget-averaged toxicity to $0.77 \pm 0.30\%$: a lower threshold admits weakly informative dimensions.


Above the band, at $\tau_z = 9$, toxicity rises again ($0.62\%$ fixed-budget, $0.69 \pm 0.27\%$ budget-averaged): a higher threshold discards dimensions that carry useful concept signal. We fix $\tau_z = 7$ for all experiments in the paper; its toxicity is essentially on par with $\tau_z = 8$, and the lower threshold retains more dimensions. 

\begin{table}[t]
\centering
\small
\resizebox{\linewidth}{!}{%
\begin{tabular}{c ccc c}
\toprule
& \multicolumn{3}{c}{Fixed budget (main-paper protocol)}
& Mean over 15 budgets \\
\cmidrule(lr){2-4}\cmidrule(lr){5-5}
$\tau_z$ & Toxicity\,\% & MMLU & PPL & Toxicity\,\% \\
\midrule
5 & 0.70 & 0.573 & 15.61 & 0.90 $\pm$ 0.43\\
6 & 0.56 & 0.572 & 15.86 & 0.77 $\pm$ 0.30 \\
7 & 0.48 & 0.569 & 16.54 & 0.57 $\pm$ 0.14 \\
8 & 0.40 & 0.590 & 16.82 & 0.52 $\pm$ 0.16 \\
9 & 0.62 & 0.597 & 16.70 & 0.69 $\pm$ 0.27 \\
\bottomrule
\end{tabular}
}
\caption{Sensitivity of DSAS+GAPS to the separability threshold $\tau_z$ on a setting (Gemma-3 4B, toxicity, layer 15). Left: operating point under the fixed capability budget of the main results (perplexity $\leq +5\%$, MMLU $\geq -3\%$; unsteered baseline). Right: best feasible toxicity averaged over $15$ capability budgets.}
\label{tab:tauz-sweep}
\end{table}

\end{document}